\documentclass[letterpaper, 10 pt, conference]{ieeeconf}  

\IEEEoverridecommandlockouts                              

\usepackage{graphicx}
\usepackage{epsfig} 
\usepackage{times} 
\usepackage{amsmath} 
\usepackage{amssymb}  
\usepackage{lipsum} 
\usepackage[dvipsnames]{xcolor}
\usepackage[caption=false, font=footnotesize]{subfig}
\usepackage[vlined,linesnumbered,ruled,norelsize]{algorithm2e}
\SetKwInOut{Input}{input}\SetKwInOut{Output}{output}
\usepackage {tikz}
\usetikzlibrary {positioning, arrows, calc, arrows.meta, patterns, shapes.geometric, bending}
\usepackage{tabularx, booktabs, multirow}
\usepackage[noadjust]{cite}
\let\labelindent\relax
\usepackage[inline]{enumitem}
\usepackage{paracol}
\usetikzlibrary{decorations.text}
\usepackage{standalone}

\title{\LARGE \bf
Adapting to Human Preferences to Lead or Follow in \\ Human-Robot Collaboration: A System Evaluation
}

\author{Ali~Noormohammadi-Asl$^{*}$, Ali~Ayub, Stephen L.\ Smith,  and Kerstin~Dautenhahn
\thanks{This research was undertaken, in part, thanks to funding from the Natural Sciences and Engineering Research Council of Canada (NSERC) and the Canada 150 Research Chairs Program.}
\thanks{The authors are with the Department of Systems Design Engineering, and the Department of Electrical and Computer Engineering at the University of Waterloo, Waterloo, ON, Canada}
\thanks{$^*${\tt\small ali.asl@uwaterloo.ca}}
}

\begin{document}

\maketitle
\thispagestyle{empty}
\pagestyle{empty}

\begin{abstract}

With the introduction of collaborative robots, humans and robots can now work together in close proximity and share the same workspace. However, this collaboration presents various challenges that need to be addressed to ensure seamless cooperation between the agents. This paper focuses on task planning for human-robot collaboration, taking into account the human's performance and their preference for following or leading. Unlike conventional task allocation methods, the proposed system allows both the robot and human to select and assign tasks to each other. Our previous studies evaluated the proposed framework in a computer simulation environment. This paper extends the research by implementing the algorithm in a real scenario where a human collaborates with a Fetch mobile manipulator robot. We briefly describe the experimental setup, procedure and implementation of the planned user study. As a first step, in this paper, we report on a system evaluation study where the experimenter enacted different possible behaviours in terms of leader/follower preferences that can occur in a user study. Results show that the robot can adapt and respond appropriately to different human agent behaviours, enacted by the experimenter. A future user study will evaluate the system with human participants.

\end{abstract}

\section{Introduction}

Collaborative robots, or cobots, represent a transition from traditional industrial robots, which work in cages isolated from humans, to robots that are able to share their workspace with humans.  This collaboration offers the opportunity to leverage the complementary capabilities of both humans and robots \cite{el2019cobot}.  Although robots are fast, accurate, precise, powerful, and capable of working for extended periods, they still lack many of the necessary skills to operate as a fully autonomous system. For instance, consider a warehouse setting where several items must be retrieved and transported to another location. While a robot can plan and navigate through different aisles and approach objects, they are not proficient enough in the pick-and-place procedure to be relied upon exclusively. Thus, systems such as \cite{dhl} and \cite{6river} have been designed where the robots guide their human coworkers to pick up locations and leave the picking task to them. This collaboration can significantly increase the system/team's performance.

In most systems designed for human-robot interaction, a human-centered design is prioritized, which focuses on accommodating human preferences. Although the same applies to human-robot collaborative systems in industry, to a certain extent, the performance of the team must also be considered to justify robot deployment. These two factors, human preference and the team's performance, do not always coincide and may even conflict. Hence, an efficient system has to be able to detect and handle these conflicting goals effectively.

In our first user study \cite{noormohammadi2021effect}, we investigated three different strategies for a simulated robot: 1- prioritizing the human's preferences, 2- prioritizing the robot (team)'s goals 3- balancing between the human preference and the team's goals. We found that the balancing strategy was able to maintain high team performance without negatively affecting the human perception of the robot or collaboration. In that study, although the human could try to lead or follow the robot through their interaction, there was no estimation or learning over the human preference, and the robot, regardless of the human's preference, applied a randomly assigned strategy. 

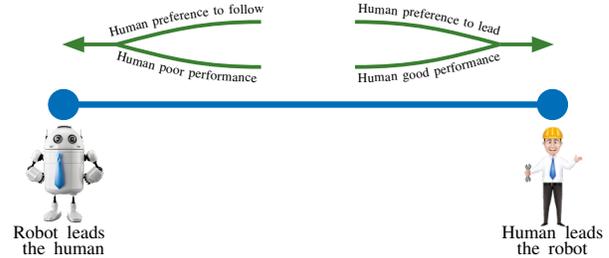
\begin{figure}[!t]
    \centering
    \begin{tikzpicture}
        \def\myshiftu#1{\raisebox{1.5ex}}
        \def\myshiftd#1{\raisebox{-2.5ex}}
        \def\myshiftu#1{\raisebox{1.5ex}}
        \def\myshiftd#1{\raisebox{-2.5ex}}
        \def\g1{7.5}
        \def\s1{1}
        \draw [line width=0.85mm, NavyBlue] (\s1, 0) -- (\g1, 0);
        \draw [color=NavyBlue, fill=NavyBlue, radius=0.2] (\s1, 0) circle[] (\g1, 0) circle[];
        \node at (\s1,-0.9) {\includegraphics[width=1.3cm]{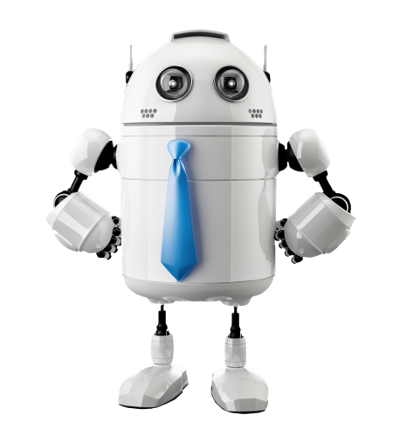}};
        \node at (\g1,-0.95) {\includegraphics[width=0.8cm]{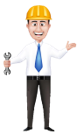}};
        
        \node at (\g1-2.75,0.8+0.3) (a1){};
        \node at (\g1-2.75,0.8-0.3) (b1){};
        \node at (\g1-0.7,0.8)[OliveGreen, line width=0.01mm, minimum size=0.0cm, align=center, inner sep=0pt] (c1){};
        \node at (\g1+0.15,0.8) (d1){};

        \node at (\s1+2.75,0.8+0.3) (a2){};
        \node at (\s1+2.75,0.8-0.3) (b2){};
        \node at (\s1+0.7,0.8)[OliveGreen, line width=0.01mm, minimum size=0.0cm, align=center, inner sep=0pt] (c2){};
        \node at (\s1-0.15,0.8) (d2){};
        
        \draw[OliveGreen,line width=0.5mm, postaction={decorate,decoration={text along path,text align=center,text={|\tiny \myshiftu|Human preference to lead}}}]  (a1) to [out=0,in=165] (c1)  ;
        \draw[OliveGreen,line width=0.5mm, postaction={decorate,decoration={text along path,text align=center,text={|\tiny \myshiftd|Human good performance}}}]  (b1) to [out=0,in=-165] (c1)  ;
        \draw[OliveGreen,-{Triangle[length=3mm,width=2mm]},line width=0.5mm]  (c1) to  (d1)  ;

         \draw[OliveGreen,line width=0.5mm, postaction={decorate,decoration={text along path,text align=center,text={|\tiny \myshiftu|Human preference to follow}}}]  (c2) to [out=15,in=180] (a2)  ;
        \draw[OliveGreen,line width=0.5mm, postaction={decorate,decoration={text along path,text align=center,text={|\tiny  \myshiftd|Human poor performance}}}]  (c2) to [out=-15,in=180] (b2)  ;
        \draw[OliveGreen,-{Triangle[length=3mm,width=2mm]},line width=0.5mm]  (c2) to  (d2)  ;
        
        \node[text width=1.5cm, align=center] at (\s1,-1.8) {\scriptsize \baselineskip=6pt Robot leads \break the human\par};
        \node[text width=1.6cm, align=center] at (\g1,-1.8) {\scriptsize \baselineskip=6pt Human leads\break the robot\par};
    \end{tikzpicture}
    \caption{The human and robot's range of leading/following roles in  human-robot collaboration}
    \label{fig:fol_lead}
\end{figure}

Based on the results obtained from the initial study, we have proposed a framework that estimates human performance and preferences for leading or following the robot and adapts itself accordingly \cite{noormohammadi2022task}. To disentangle this framework from the conventional task allocation problem, in which either one of the agents or a role allocation unit assigns tasks, in this framework, both human and robot agents have the agency to choose their own tasks, if feasible, and also assign tasks to each other. Thus, based on the human leading/following preference, the robot adapts its planning. Nevertheless, a question arises:   what if a human with a preference for leading exhibits unacceptably poor performance and thus degrades the team's performance? In the proposed framework, the robot monitors and evaluates the human agent's performance as well as the efficiency of the team and takes over the leadership role if the human agent is unable to plan efficiently (Fig.~\ref{fig:fol_lead}). To put it briefly, by default the robot is more a leader rather than a follower, unless the human prefers to have a more leading role and is able to work with acceptably high performance. 

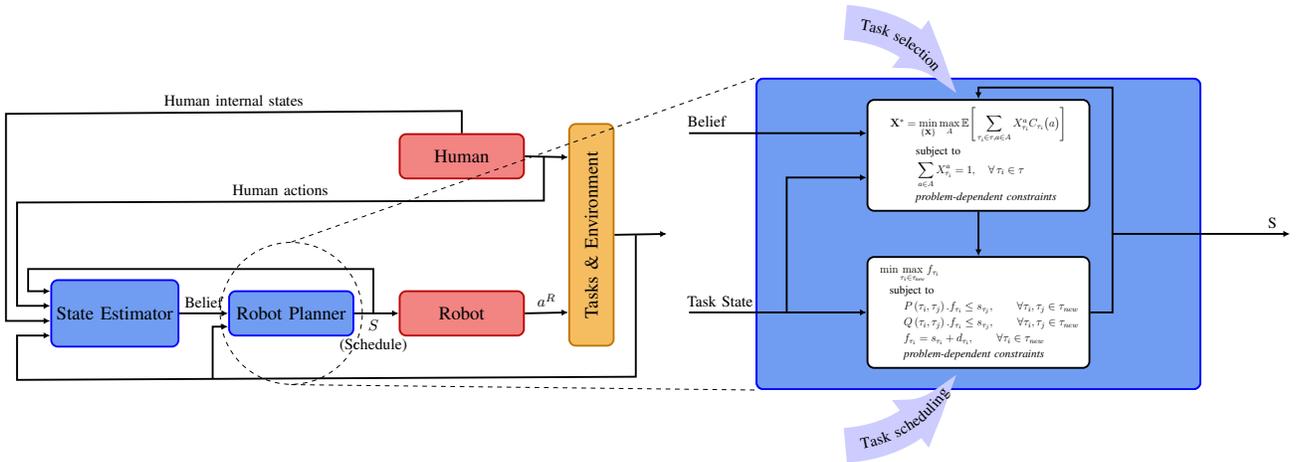
\begin{figure*}
    \centering
    \scalebox{0.59}{
    \begin{tikzpicture}[%
    ,node distance=10mm
    ,>=latex'
    ,every path/.style={->} ,
    mypostaction/.style n args=3{
   decoration={
       text align={
           left indent=#1},
       text along path, 
       text={#2},#3
   },
   decorate
   }
    ]
    \tikzstyle{arrow1} = [-{Latex[length=1.5mm,width=1.5mm]}, very thick];
    \tikzstyle{block1} = [draw ,minimum height=10mm,minimum width=28mm ,align=center, rounded corners=.15cm, font=\large, very thick];
    \tikzstyle{output} = [coordinate];
    \definecolor{cl1}{HTML}{F28482};
    \definecolor{cl11}{HTML}{BB0704};
    \definecolor{cl2}{HTML}{F6BD60};
    \definecolor{cl22}{HTML}{BF7600};
    \definecolor{cl3}{HTML}{709DF1};
    
    \node [block1, minimum height = 1.5cm, fill=cl3, draw=blue] (a) {State Estimator};
    \node [block1, right= 1.1cm of a, fill=cl3, draw=blue] (b) {Robot Planner};
    \node [block1, right=of b, fill=cl1, draw=cl11] (c) {Robot};
    \node [block1, above=of c, yshift=1.5cm, fill=cl1, draw=cl11] (d) {Human};
    \node [block1, minimum width=5cm, right=of c, xshift=0.5cm, yshift=-0.75cm, rotate= 90, fill=cl2, draw=cl22] (e) {Tasks \& Environment};
    \node [output, right=of e, xshift=0.7cm, yshift=-2.5cm] (o) {};
    \node [inner sep=0,minimum size=0,right of=b, xshift= 0.85cm, label={[label distance=-1cm, align=center]90:{$S$\\(Schedule)}}] (p) {}; 
    \node [inner sep=0,minimum size=0,right of=d, xshift= 0.85cm] (u) {}; 
    \node [inner sep=0,minimum size=0,right of=e] (k) {}; 
    \draw [arrow1] (a) to node[above] {Belief} (b);
    \draw [arrow1] (b) to (c);
    \draw [arrow1] (c.east) to node[above] {$a^R$} ($(e.north)+(0,-1.75)$) ;
    \draw [arrow1] (d.east) to ($(e.north)+(0,1.75)$);
    \draw [arrow1] (d.north) -- +(0,0.5) -- node[above] {Human internal states} ($(a.west)+(-1 ,4.5)$) -- ($(a.west)+(-1 ,0)$) |- ($(a.west)+(0,-0.17)$);
    \draw [arrow1] (k) -- +(0,-3.2) -- ($(a.west)+(-0.75,-1.5)$) |- ($(a.west)+(0,-0.5)$);
    \draw [arrow1] (e.south) -- (o.west);
    \draw [arrow1] ($(b.west)+(-0.35,-1.48)$) |- ($(b.west)+(0,-0.3)$);
    \draw [arrow1] (p) -- +(0,1) -- ($(a.west)+(-0.5,1)$) |- ($(a.west)+(0,0.5)$);
    \draw [arrow1] (u) -- +(0,-1) -- node[above] {Human actions} ($(a.west)+(-0.75,2.5)$) |- ($(a.west)+(0,0.17)$);
    \node [block1, right = 2cm of o,minimum width=10cm, minimum height = 7cm, fill=cl3, draw=blue] (pl) {};
    \node [block1, right = 2cm of o,minimum width=10cm, minimum height = 7cm, fill=cl3, draw=blue] (pl) {};
    \node [block1, above = 0.5cm of pl.center,minimum width=5cm, minimum height = 2.5cm, draw=black, fill=white] (so) {\scalebox{0.6}{$\begin{aligned}&\mathbf{X}^* = \min_{\left\{\mathbf{X}\right\}}\max_{A} \mathbb{E} \Bigg[\sum_{\tau_i\in \tau, a \in A}X_{\tau_i}^a C_{\tau_i}\big(  a\big)\Bigg] \\
    &\qquad \; \text{subject to}\\
    &\qquad \; \sum_{a\in A}{X_{\tau_i}^a}=1, \quad \forall \,\tau_i\in \tau \\
    &\qquad \; \textit{problem-dependent constraints}
    \end{aligned}$}};
    \node [block1, below = 0.5cm of pl.center,minimum width=5cm, minimum height = 2.5cm, draw=black, fill=white] (do) {\scalebox{0.6}{
        $\begin{aligned}
            &\min\max_{\tau_i \in \tau_{\textit{new}}} f_{\tau_i}\\ 
            &\quad\text{subject to}\\
            &\qquad \; P\left(\tau_{i},\tau_{j}\right).f_{\tau_{i}}\leq s_{\tau_{j}}, \qquad  \forall \tau_{i},\tau_{j}\in \tau_{new}\\
            &\qquad \; Q\left(\tau_{i},\tau_{j}\right).f_{\tau_{i}}\leq s_{\tau_{j}}, \qquad  \forall \tau_{i},\tau_{j} \in \tau_{new}\\
            &\qquad \; f_{\tau_i} = s_{\tau_i} + d_{\tau_i}, \qquad \forall \tau_i\in \tau_{new} \\
            &\qquad \; \textit{problem-dependent constraints}
        \end{aligned}$
    }};
     \draw [arrow1] ($(so.west)+(-1.8,-3.52)$) |- ($(so.west)+(0,-0.5)$);
     \draw [arrow1] ($(so.west)+(-4,0.5)$) node[above, xshift=0.4cm] {Belief} -- ($(so.west)+(0,0.5)$);
     \draw [arrow1] ($(do.west)+(-4,0)$) |- node[above, xshift=0.7cm] {Task State} ($(do.west)$);
     \draw [arrow1] ($(so.south)$) -- ($(do.north)$);
     \draw[arrow1] (do.east) -| ($(pl.east)+(-2, 0)$) -- ($(pl.east)+(2, 0)$) node[above, xshift=-0.4cm] {S};
     \draw[arrow1] ($(pl.east)+(-2, 0)$) -- ($(pl.east)+(-2, 3.3)$) -| (so.north);
    \node [draw, circle, dashed, below=0mm of b.center, anchor=center, minimum size=32mm] (zp){};
     \path[dashed](zp.south) edge[-] (pl.south west){};
     \path (zp.north) edge[-, dashed] (pl.north west){};

     \draw[-{Latex[bend,length=5em]}, blue!20!white, line width=5ex,postaction={mypostaction={1em}{Task selection}{raise=-0.7ex}}] (so)++ (-3,3)  to[bend left]  ($(so.north)+ (-0.5,0)$);
      \draw[-{Latex[bend,length=5em]}, blue!20!white, line width=5ex,postaction={mypostaction={1em}{Task scheduling}{raise=-0.7ex}}] (do)++ (-3,-3)  to[bend right]  ($(do.south)+ (-0.5,0)$);
    \end{tikzpicture} }

    \caption{Task selection and planning architecture}
    \label{fig:architect}
\end{figure*}

This approach moves the collaborative system beyond a strict and explicit leading/following structure and instead covers the entire range between the two ends of this spectrum: being a leader or a follower. Fig.~\ref{fig:fol_lead} illustrates the general idea of this framework. In \cite{noormohammadi2022task}, we elaborated this framework and showed its proficiency through a computer simulation study, where the human and robot needed to collaborate to finish a task.

Despite the satisfactory results of the proposed framework in our previous work, the effectiveness of a human-related framework can only be fully verified when it is applied in the presence of real human participants, with all the possible uncertainties arising due to their presence. Hence, we have designed and implemented a collaborative scenario, in which the human works alongside a cobot, Fetch \cite{wise2016fetch}. As a first step, we carry out a system evaluation where the experimenter takes the roles of human participants with different leading/following preferences. The goal was to  assess the effectiveness of the proposed framework in a real-world situation, to allow us to finalize the system before starting a future user study.

\subsection{Related work}

The topic of task scheduling and allocation in human-robot collaboration has garnered interest in recent cobot research. The focus has been on optimizing task allocation and efficient scheduling by leveraging the capabilities of both humans and robots. While our work differs from conventional task scheduling problems due to the agency of both humans and robots in selecting and assigning tasks, similarities between the two make it valuable to examine prior research in this area. Moreover, reviewing studies on human-robot one-sided or mutual adaptation can provide insights to bridge the gap between human-robot adaptation and the task scheduling problem. 


\subsubsection{Task scheduling} Task allocation and scheduling problems for multi-robot systems have been studied extensively\cite{khamis2015multi, korsah2013comprehensive}. In some studies, the same approaches for multi-robot systems have been applied to human-robot systems with the difference that a robot has been replaced with a human agent with different capabilities. In these studies, usually,  offline scheduling methods have been proposed with the goal of minimizing the task completion cost and time \cite{tsarouchi2017human, lee2022task}. However, due to the uncertainty arising from the human agents' presence, offline scheduling methods cannot be an effective approach in most cases. Thus, some  task scheduling frameworks with dynamic and online planning abilities have been proposed \cite{cheng2021human, casalino2019optimal, fusaro2021integrated, zhang2020real}.

Another important and often overlooked factor, which is the focus of our work, is considering the human preference in task allocation and scheduling, which has been shown to be a key factor in human satisfaction and positive perception of the robot. In \cite{tausch2020best}  different scenarios have been explored in which tasks were assigned by the manager, the robot, or the participants themselves. Similarly, researchers in \cite{gombolay2017computational} asked participants about their preference for particular tasks and considered it in the offline planning. However, these studies have considered the human agents' preferences during the offline planning phase and by asking them directly. More importantly, they do not consider participants' performance to check if their preference is aligned with the team's performance or deteriorating it.

To have fluid and successful human-robot collaboration, adaptation to the other agent's preference is crucial. One approach to achieving this is to have experts and developers manually instruct the robot on how to complete tasks while taking human preferences into account. Although this method is fast and effective, is not practical in many applications since it is \cite{unhelkar2020effective, hiatt2017human}:
\begin{enumerate*}
    \item  time-consuming, as the expert has to consider different possible actions and scenarios,
    \item  not expressive, as it is not possible to evaluate all aspects and scenarios,
    \item  not scalable, since new designs and modeling are needed upon a change.
\end{enumerate*}

Supervised and unsupervised learning-based methods have been employed to learn the human agent's preference in real-time. In supervised learning-based approaches, such as k-nearest neighbors \cite{admoni2014data}, neural networks and deep learning \cite{soh2020multi}, and support vector machines \cite{castellano2012detecting}, the human's preference is learned by identifying decision factors for data collected on human behaviour and enriching them through annotation and surveying. Additionally, supervised learning-based methods have been applied to modeling human behaviours and preferences after observing their behaviour \cite{nikolaidis2017human, luo2018unsupervised, nikolaidis2015efficient}.
However, these studies mostly focus on the robot's adaptation to the human agent. To consider both the human's preference and performance in a collaborative scenario, mutual adaptation enabling the human and robot to adjust their behaviour and performance based on each other's actions and feedback is needed.
The authors of \cite{nikolaidis2017human} explored the concept of mutual adaptation between human and robot, which allows the robot to guide an adaptable human, rather than adapting to the human. However, in situations where the human agent's performance is inadequate, the robot eventually ends up adapting itself one-sidedly.

\subsection{Contributions}
There are three noteworthy contributions in our work:
\begin{enumerate*}
    \item We designed a real scenario where the human and robot collaborate to accomplish a task with the agency to select their own tasks or assign tasks to each other.
    \item We implemented our framework on a real, mobile manipulator robot empowering it to adapt its leading/following role based on the human's performance and preference.  
    \item We show that the robot can successfully adapt its behaviour to a variety of possible human behaviours, from high/low-performing humans with a strong/weak desire to lead, to humans whose performance degrades over time.
\end{enumerate*}
\section{Planning Strategy}\label{sec:2}

This section explains the problem statement and the planning strategy briefly. A more detailed discussion can be found in \cite{noormohammadi2022task}.

\subsection{Problem Statement}
In this work, we consider a single-human single-robot collaboration, where the team needs to finish a set of precedence-constrained tasks successfully. Due to their differing abilities, the required time for completing the same task can be different for the human and robot. If we assume the collaboration time as the objective function,  in conventional task allocation approaches, tasks are assigned to minimize collaboration time without considering the human's preferences. To address this limitation, in our work, the agents have the agency to choose their own tasks and to assign tasks to their coworker at each decision step. Therefore, at each step, the robot needs to 
\begin{enumerate*}
    \item estimate the human's leading/following preference
    \item monitor and estimate the human's performance level
    \item  Select and assign tasks to minimize the collaboration cost, 
    \item and finally, detect and correct human errors.
\end{enumerate*}

\subsection{Planning Architecture}
The planning architecture is depicted in Fig.~\ref{fig:architect}. The state estimator has a key role in this architecture, as the robot needs to estimate the human's preference and performance based on the history of the human and robot's actions, the robot's previous beliefs, the human's internal states (e.g., speed and fatigue), and the tasks' states. The output of this unit, along with the tasks and environment's states, are fed into the robot planner to determine the robot's next actions. In this unit, the robot first selects tasks that are needed to be done by itself and the human by solving a stochastic optimization problem. Then, it solves a deterministic optimization problem for scheduling the selected tasks. Finally, the robot performs its next actions according to the output schedule. The optimization problems shown in the robot planner unit (Fig.~\ref{fig:architect}) are elaborated in \cite{noormohammadi2022task}.

\section{Envisaged User Study Setup}

\begin{figure*} 
    \centering
  \subfloat[Main pattern]{%
       \includegraphics[width=0.45\linewidth]{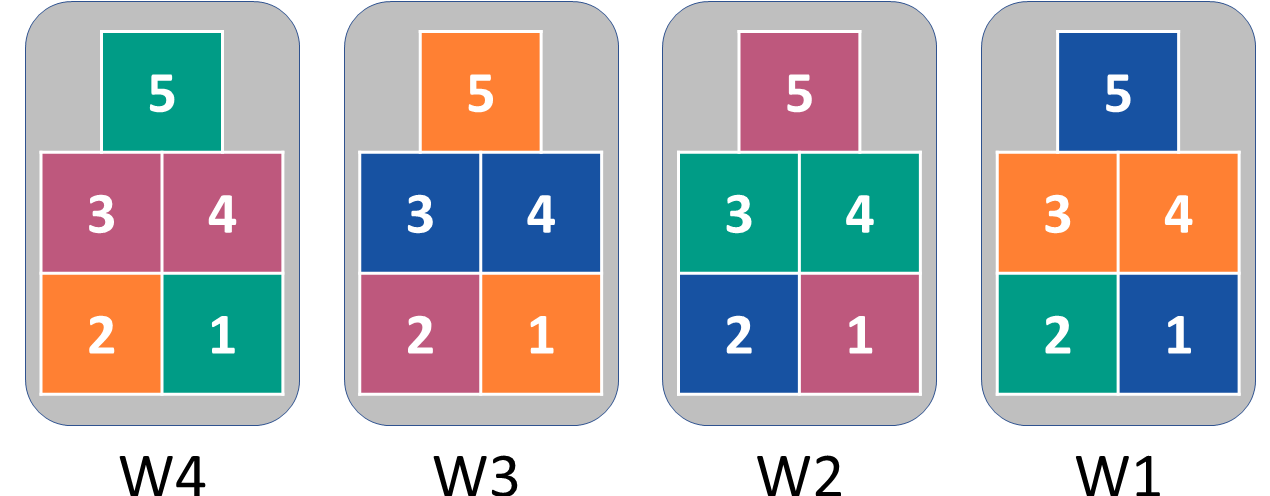}}
    \hfill
  \subfloat[Easy]{%
        \includegraphics[width=0.45\linewidth]{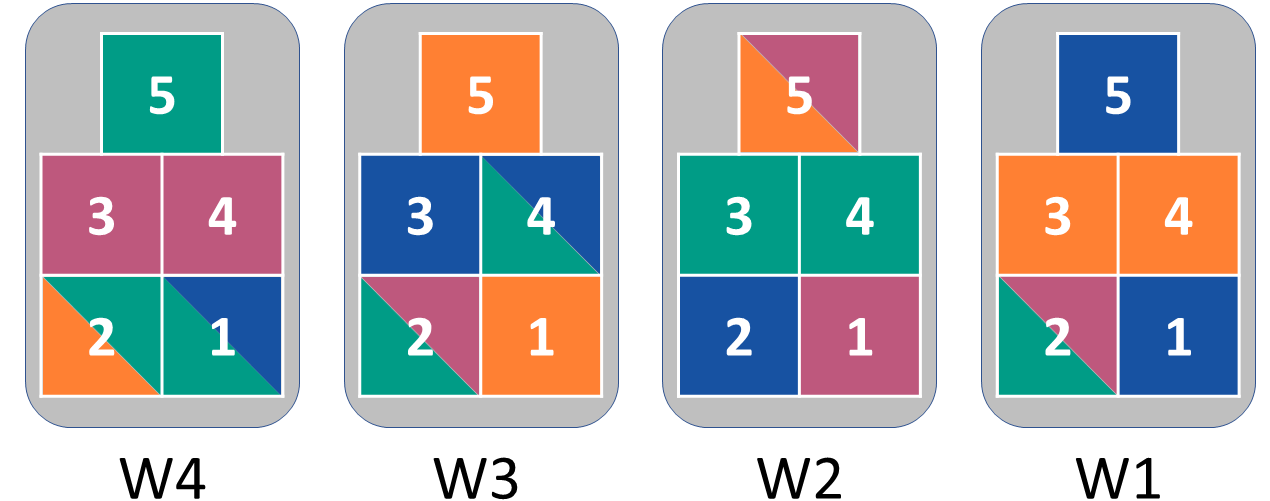}}
    \\
  \subfloat[Medium]{%
        \includegraphics[width=0.45\linewidth]{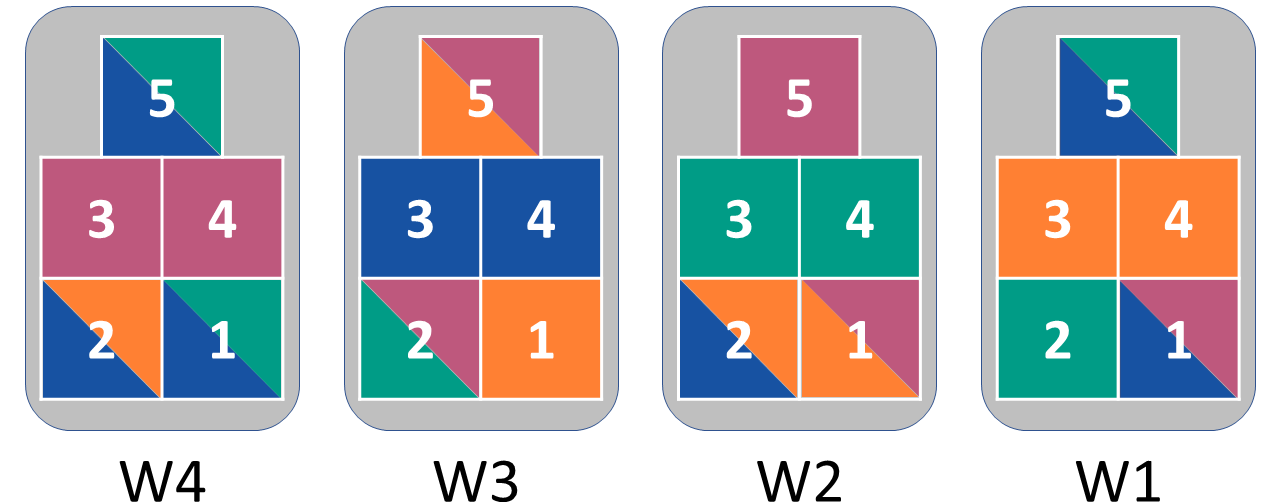}}
    \hfill
  \subfloat[Difficult]{%
        \includegraphics[width=0.45\linewidth]{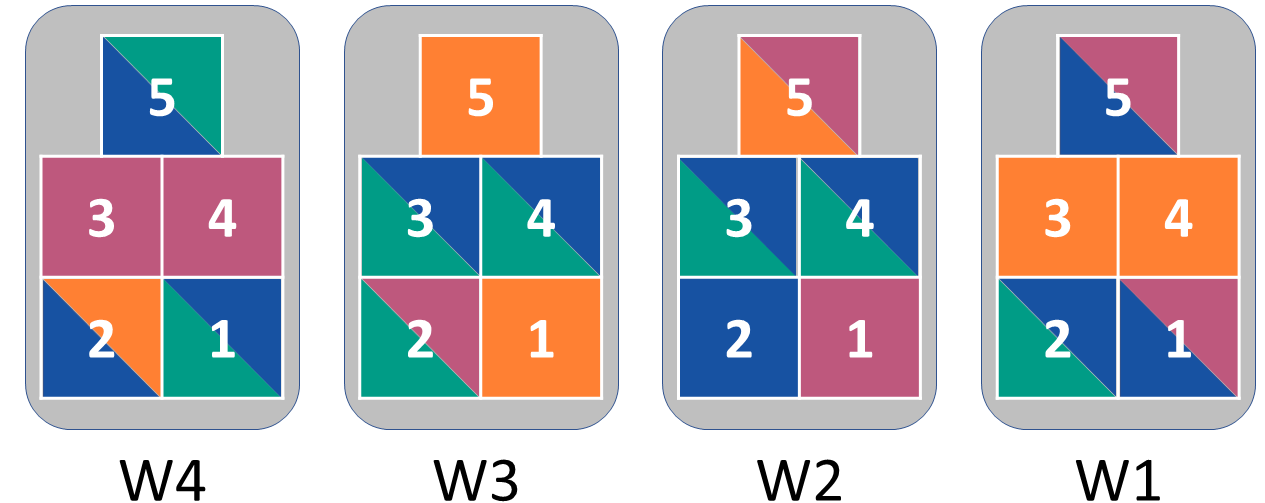}}
  \caption{(a) The main patterns shown at the start. (b) An easy, (c) a medium, and (d) a difficult variant of the main patterns with some partially unknown color of the spots}
  \label{fig:patterns} 
\end{figure*}

In this section, to provide the context for the present paper, we describe the procedure of the future user study involving a human-robot team that must place a specified number of colored objects onto shared workspaces according to a predetermined pattern. Objects are in four different colors: green, blue, pink, and orange. Fig.~\ref{fig:exp_outline} illustrates the experiment environment, which includes four workspaces in the shared area, each with five numbered spots. On each workspace, participants have to follow the order of the numbers when filling the spots. At the start of the experiment, participants will receive a set of colored patterns for the four workspaces and will have a given amount of time to memorize them. Next, the pattern sheet is collected, and participants will be given the same patterns again but with partially available information, in which some spots are filled with two colors, one of which is correct. The level of difficulty for the task is determined by the amount of available information on the new pattern sheet; the more information is available (i.e., known pattern), the easier the task. Fig.~\ref{fig:patterns} shows a sample pattern and its partially available versions with different difficulty levels: easy, medium, and difficult. The reasoning behind choosing this pattern memorization task is to make it cognitively challenging for a short collaboration scenario, as it is not feasible to conduct a long experiment that could leave participants mentally and physically exhausted. Therefore,  relatively short tasks (each about 12-20 minutes) that are still mentally stimulating were chosen.

The human's objects are distributed across two tables: one (Table 1 in Fig.~\ref{fig:exp_outline}) is located far from the workspace (placing locations, Table 1), and the other table (Table 3) is located close to the workspace. On the closer table, there are green and orange objects, while blue and pink objects are on the farther away table. Similarly, the robot's objects are also located on two tables, either far and close to the shared space. The farther away table contains blue and green objects, the closer table contains pink and orange objects. Clearly, the agents have to travel a longer distance when needing an object on the farther away table. The rationale behind this distribution is to have
\begin{enumerate*}
    \item a set of  objects (blue blocks) far from both;
    \item a set of objects (green blocks) close to both;
    \item a set of objects (green blocks) far from the human but close to the robot; and
    \item a set of objects (pink blocks) close to the human but far from the robot.
\end{enumerate*}

The purpose of this object distribution is to evaluate the human agent's decision-making in terms of assigned load distribution and travel distance, and its proximity to the optimal solution for task completion time. By strategically placing objects on two tables at different distances from the workspace, we can assess how effectively the human agent allocates tasks and navigates the environment to complete the tasks in the most efficient manner.

This distribution is expected to help assess the decision-making  of the human agent in terms of task selection and assignment to the robot by evaluating the distribution of the assigned load (travel distance) and its proximity to the optimal solution for task completion time.

To facilitate effective communication between the human and robot, a graphical user interface (GUI) was designed, installed on a tablet computer, and placed on the human's table, which is near the shared workspace. Through this GUI, the human and robot can assign tasks to each other and inform each other which action they are currently performing. During each action, the human and robot are allowed to fetch and place only one object. The GUI prevents the human agent from taking infeasible actions like violating precedence constraints or selecting a task the robot is already performing.  It is essential for the agents to inform each other when they start and finish a task. Table \ref{tab:actions} shows the set of defined actions for agents. The robot's autonomy to reject the task assigned by the human (action 6 in Table \ref{tab:actions}), if necessary, offers the robot the agency to adjust its leading role in the collaboration.

\begin{table*}[]
    \centering
    \begin{tabular}{@{}l|l@{}}
    \toprule
        \multicolumn{1}{c}{\textbf{Human}} & \multicolumn{1}{c}{\textbf{Robot}} \\ \midrule
        1- Selecting a task for themselves  & 1- Selecting a task for itself \\
        2- Assigning a task to the robot  & 2- Assigning a task to the human\\
        3- Returning an object from the shared workspace & 3- Returning a wrong object from the shared workspace \\
        4- Performing a task assigned by the robot & 4- Performing a correct task assigned by the human \\
        5- Canceling a task assigned to the robot & 5- Canceling a Task assigned to the human \\
        \multicolumn{1}{c|}{---} & 6- Rejecting a task assigned by the human\\ \bottomrule
    \end{tabular}
    \caption{The human and robot's sets of possible actions}
    \label{tab:actions}
\end{table*}

When the robot removes a misplaced object by the human from the shared workspace, for safety considerations, the robot releases it on top of a conveyor belt that is transferring the object from the robot's side of the table to the human's side of the table. This keeps the human's and robot's workspace separate. In addition, again, for safety considerations, we have installed a smart light bulb on top of the shared workspace that turns red when the robot places or picks an object on/from the table, signalling the human not to approach the shared area for picking or placing a block. We have also completely separated the two agents' workspaces with safety tapes and cones placed on the floor. In addition to these considerations, the experimenters will constantly monitor the robot to intervene by either controlling the robot with the joystick or stopping the robot using the safety stop button.

\subsection{Robot}
The mobile manipulator robot used in this experiment is the Fetch robot \cite{wise2016fetch}. We have equipped the robot with autonomous pick-and-place functionality that allows it to
\begin{enumerate}
\item navigate to the designated pick-up location,
\item use its RGB-D camera to detect and locate markers on the blocks or tables,
\item store and update marker locations in its memory,
\item plan a motion for its arm to pick up the object,
\item move to the designated place location, and
\item plan a motion for placing the object and releasing it once the robot detects contact with the table's top surface.
\end{enumerate}
For the sake of brevity, we have omitted the technical details of this robotics implementation.

\subsection{Procedure of Future User Study}

We designed and implemented the experiment in view of  a future  user study with human participants, to be summarized here briefly in order to explain the rationale behind the system evaluation reported in this paper. The envisaged user study will consist of two phases: an in-person phase and an online phase. During the in-person phase, participants will collaborate with the robot and complete a set of questionnaires. In the online phase, participants will watch a recorded video of their interaction with the robot and explain the reasoning behind their actions. They will also be asked to complete two questionnaires about their leadership and followership styles, as well as provide an open-ended response about their ideas for improving the collaboration. However, as this paper focuses primarily on the proposed framework and system evaluation, rather than a user study, we will not report questionnaire data.

During the in-person phase, we will first introduce participants to the task and explain how to use the GUI. To achieve this, we will ask them to perform a medium-difficulty pattern without the robot's help. This serves a dual purpose: first, it will familiarize participants with the task and GUI, and second, it will provide us with an estimate of their speed and performance.
Next, we will provide an easy pattern for the human to perform with the assistance of the robot. The human will initiate the task, and the robot will  begin planning and taking action as soon as the human starts placing an object on the table. This will allow the human to assign tasks to the robot at the beginning of the experiment and will provide the robot with a better understanding of the human's preferences to lead. In the third and fourth parts of the in-person phase, we will present the human with patterns of medium and hard difficulty levels, respectively. 
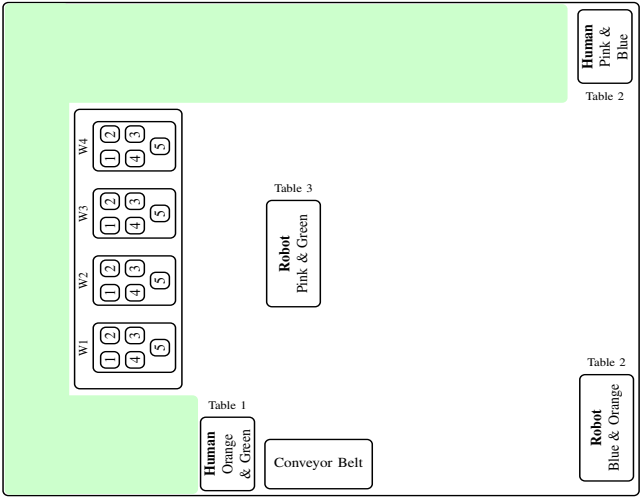
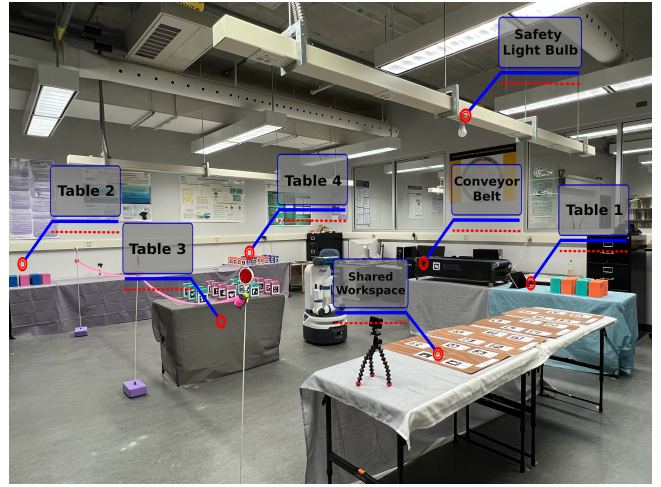
\begin{figure*}
    \centering
    \subfloat[Outline of the experiment environment\label{fig:exp_outline}]{%
        \begin{tikzpicture}[%
    ,node distance=10mm
    ,>=latex'
    ,every path/.style={->} ,
    mypostaction/.style n args=3{
   decoration={
       text align={
           left indent=#1},
       text along path, 
       text={#2},#3
   },
   decorate
   }
    ]
    \tikzstyle{arrow1} = [-{Latex[length=1.5mm,width=1.5mm]}, very thick];
    \tikzstyle{block1} = [draw ,minimum height=10mm,minimum width=28mm ,align=center, rounded corners=.15cm, font=\large, very thick];
    \tikzstyle{output} = [coordinate];
    \definecolor{cl1}{HTML}{F28482};
    \definecolor{cl11}{HTML}{BB0704};
    \definecolor{cl2}{HTML}{F6BD60};
    \definecolor{cl22}{HTML}{BF7600};
    \definecolor{cl3}{HTML}{709DF1};

        \node[block1, minimum width = \linewidth, minimum height=15cm] (f) {};
        \node[block1, right= 2cm of f.west, minimum width = 3cm, minimum height=8.5cm] (sw){};
        \node[block1, above right= 0.5cm and -1.0 of sw.south, minimum width = 2.3cm, minimum height=1.5cm](w1){};
            \node[left= 0.01cm of w1.west]{\rotatebox{90}{W1}};
            \node[block1, below left= 0.1cm and 0.4cm of w1.center, minimum width = 0.5cm, minimum height=0.5cm](w11){\rotatebox{90}{1}};
            \node[block1, above left= 0.1cm and 0.4cm of w1.center, minimum width = 0.5cm, minimum height=0.5cm](w12){\rotatebox{90}{2}};
            \node[block1, above left= 0.1cm and -0.3cm of w1.center, minimum width = 0.5cm, minimum height=0.5cm](w13){\rotatebox{90}{3}};
            \node[block1, below left= 0.1cm and -0.3cm of w1.center, minimum width = 0.5cm, minimum height=0.5cm](w14){\rotatebox{90}{4}};
            \node[block1, left= -1cm of w1.center, minimum width = 0.5cm, minimum height=0.5cm](w15){\rotatebox{90}{5}};
        \node[block1, above= 0.5cm of w1.north, minimum width = 2.3cm, minimum height=1.5cm](w2){};
            \node[left= 0.01cm of w2.west]{\rotatebox{90}{W2}};
            \node[block1, below left= 0.1cm and 0.4cm of w2.center, minimum width = 0.5cm, minimum height=0.5cm](w21){\rotatebox{90}{1}};
            \node[block1, above left= 0.1cm and 0.4cm of w2.center, minimum width = 0.5cm, minimum height=0.5cm](w22){\rotatebox{90}{2}};
            \node[block1, above left= 0.1cm and -0.3cm of w2.center, minimum width = 0.5cm, minimum height=0.5cm](w23){\rotatebox{90}{3}};
            \node[block1, below left= 0.1cm and -0.3cm of w2.center, minimum width = 0.5cm, minimum height=0.5cm](w24){\rotatebox{90}{4}};
            \node[block1, left= -1cm of w2.center, minimum width = 0.5cm, minimum height=0.5cm](w25){\rotatebox{90}{5}};
        \node[block1, above= 0.5cm of w2.north, minimum width = 2.3cm, minimum height=1.5cm](w3){};
            \node[left= 0.01cm of w3.west]{\rotatebox{90}{W3}};
            \node[block1, below left= 0.1cm and 0.4cm of w3.center, minimum width = 0.5cm, minimum height=0.5cm](w31){\rotatebox{90}{1}};
            \node[block1, above left= 0.1cm and 0.4cm of w3.center, minimum width = 0.5cm, minimum height=0.5cm](w32){\rotatebox{90}{2}};
            \node[block1, above left= 0.1cm and -0.3cm of w3.center, minimum width = 0.5cm, minimum height=0.5cm](w33){\rotatebox{90}{3}};
            \node[block1, below left= 0.1cm and -0.3cm of w3.center, minimum width = 0.5cm, minimum height=0.5cm](w34){\rotatebox{90}{4}};
            \node[block1, left= -1cm of w3.center, minimum width = 0.5cm, minimum height=0.5cm](w35){\rotatebox{90}{5}};
        \node[block1, above= 0.5cm of w3.north, minimum width = 2.3cm, minimum height=1.5cm](w4){};
            \node[left= 0.01cm of w4.west]{\rotatebox{90}{W4}};
            \node[block1, below left= 0.1cm and 0.4cm of w4.center, minimum width = 0.5cm, minimum height=0.5cm](w41){\rotatebox{90}{1}};
            \node[block1, above left= 0.1cm and 0.4cm of w4.center, minimum width = 0.5cm, minimum height=0.5cm](w42){\rotatebox{90}{2}};
            \node[block1, above left= 0.1cm and -0.3cm of w4.center, minimum width = 0.5cm, minimum height=0.5cm](w43){\rotatebox{90}{3}};
            \node[block1, below left= 0.1cm and -0.3cm of w4.center, minimum width = 0.5cm, minimum height=0.5cm](w44){\rotatebox{90}{4}};
            \node[block1, left= -1cm of w4.center, minimum width = 0.5cm, minimum height=0.5cm](w35){\rotatebox{90}{5}};

        \node[block1, below right= 3.1cm and 2cm of sw.south, minimum width = 2cm, minimum height=1.5cm, text width=2cm,align=center, rotate=90] (h1){\textbf{Human}\\ Orange \& Green}; \node[above=1.25cm of h1.center]{Table 1};
        \node[block1, above left= 7.3cm and 0.2cm of f.east, minimum width = 1.5cm, minimum height=1.5cm, text width=2cm,align=center, rotate=90] (h2){\textbf{Human}\\ Pink \& Blue};\node[below=1.25cm of h2.center]{Table 2};
        \node[block1, above left= 1.5cm and 0.0 of f.center, minimum width = 2cm, minimum height=1.5cm, rotate=90, text width=3cm,align=center] (r1){\textbf{Robot} \\ Pink \& Green};\node[above=1.75cm of r1.center]{Table 3};
        \node[block1, below left= 3.8cm and 1.7cm of f.east, minimum width = 2cm, minimum height=1.5cm, text width=3cm,align=center, rotate=90] (r2){\textbf{Robot} \\ Blue \& Orange};\node[above=1.75cm of r2.center]{Table 2};
        \node[block1, below right= 1.5cm and 3.8 cm of sw.south, minimum width = 3cm, minimum height=1.5cm] (sw){Conveyor Belt};

        \node[block1,draw=none, below left= 0.05cm and 2cm of f.north east, minimum width = 15.5cm, minimum height=3cm, fill=green!20!white] (hs1) {};
        \node[block1,draw=none, right= .05cm of f.west, minimum width = 1.8cm, minimum height=14.95cm, fill=green!20!white] (hs2) {};
        \node[block1,draw=none, above right= 0.05cm and 0.05cm of f.south west, minimum width = 5.4cm, minimum height=3cm, fill=green!20!white] (hs2) {};
    \end{tikzpicture}
      }
      \hfill
    \subfloat[Experimental setup\label{fig:exp_setup}]{%
        \includegraphics[width=0.48\textwidth]{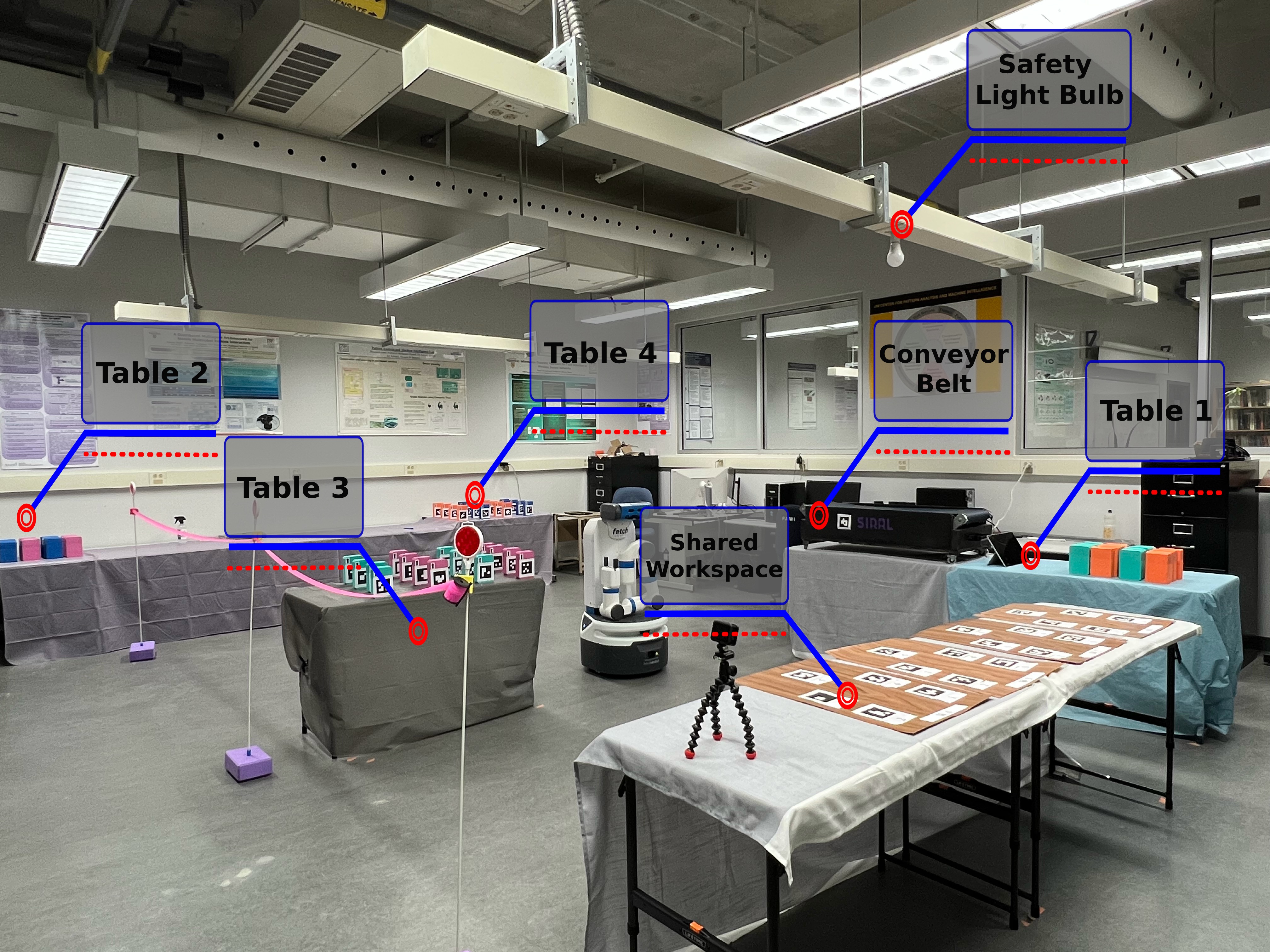}
      }
    \caption{Experiment Environment}
    \label{fig:experiment_env}
\end{figure*}

\subsection{Procedure of System Evaluation}
Concerning the explained future user study, we performed a   system evaluation study prior to the user study experience to assess the algorithm performance, identify issues, optimize the system, improve the user experience, and validate the system. This can ultimately lead to more accurate and reliable results in our user study.

To do so, we mainly focused on the in-person phase, where participants need to collaborate with the robot, and enacted four possible different behaviours that participants may have during the experiment. These behaviours were intentionally enacted by the experimenter and included:
\begin{enumerate*}
\item a preference for following with low accuracy
\item a preference for following with high accuracy
\item a preference for leading with high accuracy
\item a preference for leading with low accuracy
\end{enumerate*}

We also took into account the scenario that the participant preferred to be the leader and performed well initially, but then forgot the pattern and made errors while still wanting to lead the robot. By considering this case, we aimed to assess the robot's ability to adapt quickly to new conditions. Furthermore, this scenario represents a possible situation where participants remember the pattern better in the beginning but gradually forget it over time. After outlining the system evaluation goal and study cases, we will now move on to explaining the experiment environment and procedure.

\begin{figure*}
    \centering
    \subfloat[Table1: The human is working with the GUI. Green and orange objects are on this table\label{fig:d1}]{%
        \includegraphics[width=0.23\textwidth]{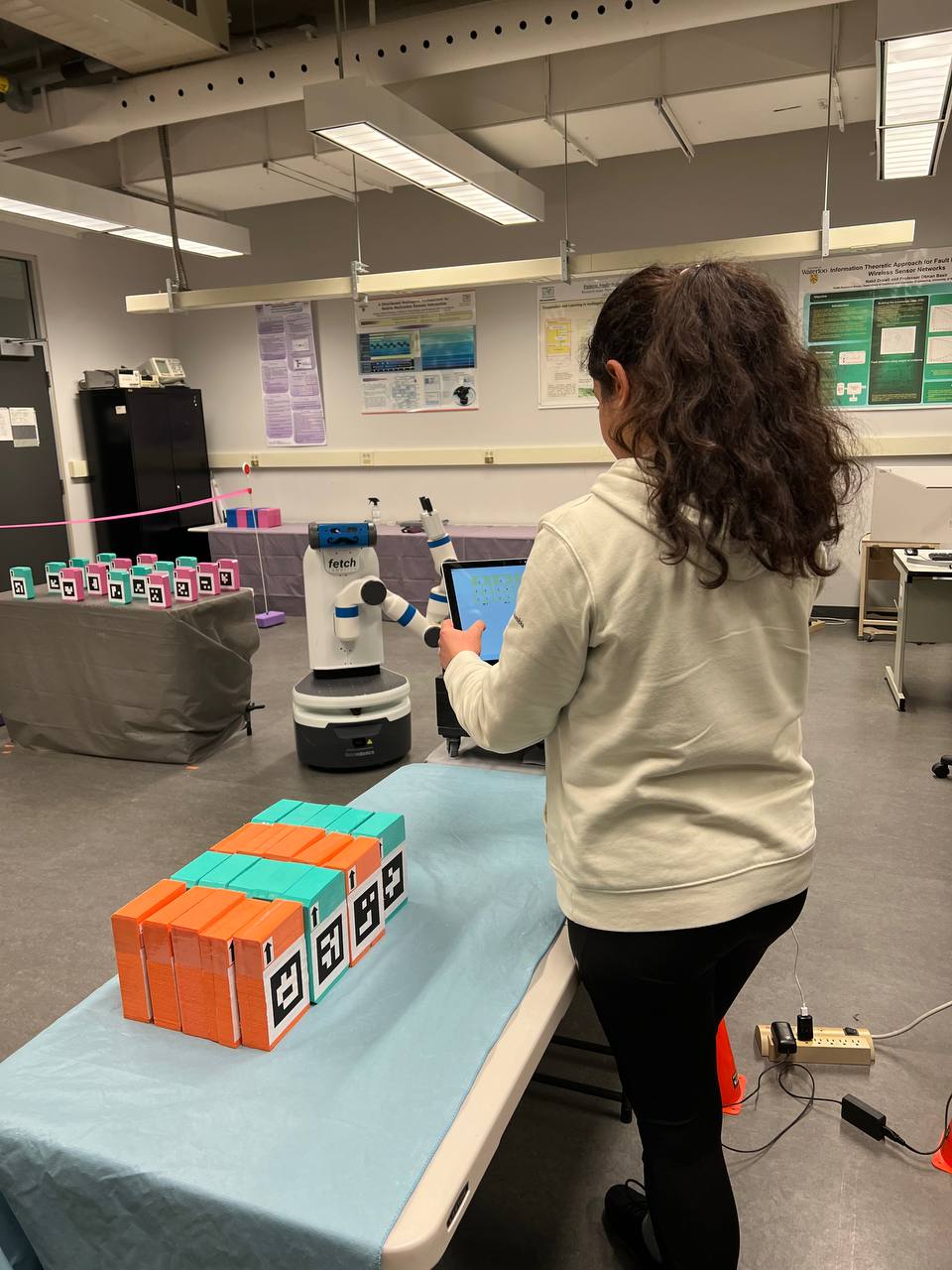}
      }
    \hfill
    \subfloat[Fetch is ready to start  when the human selects her first task\label{fig:d2}]{%
        \includegraphics[width=0.23\textwidth]{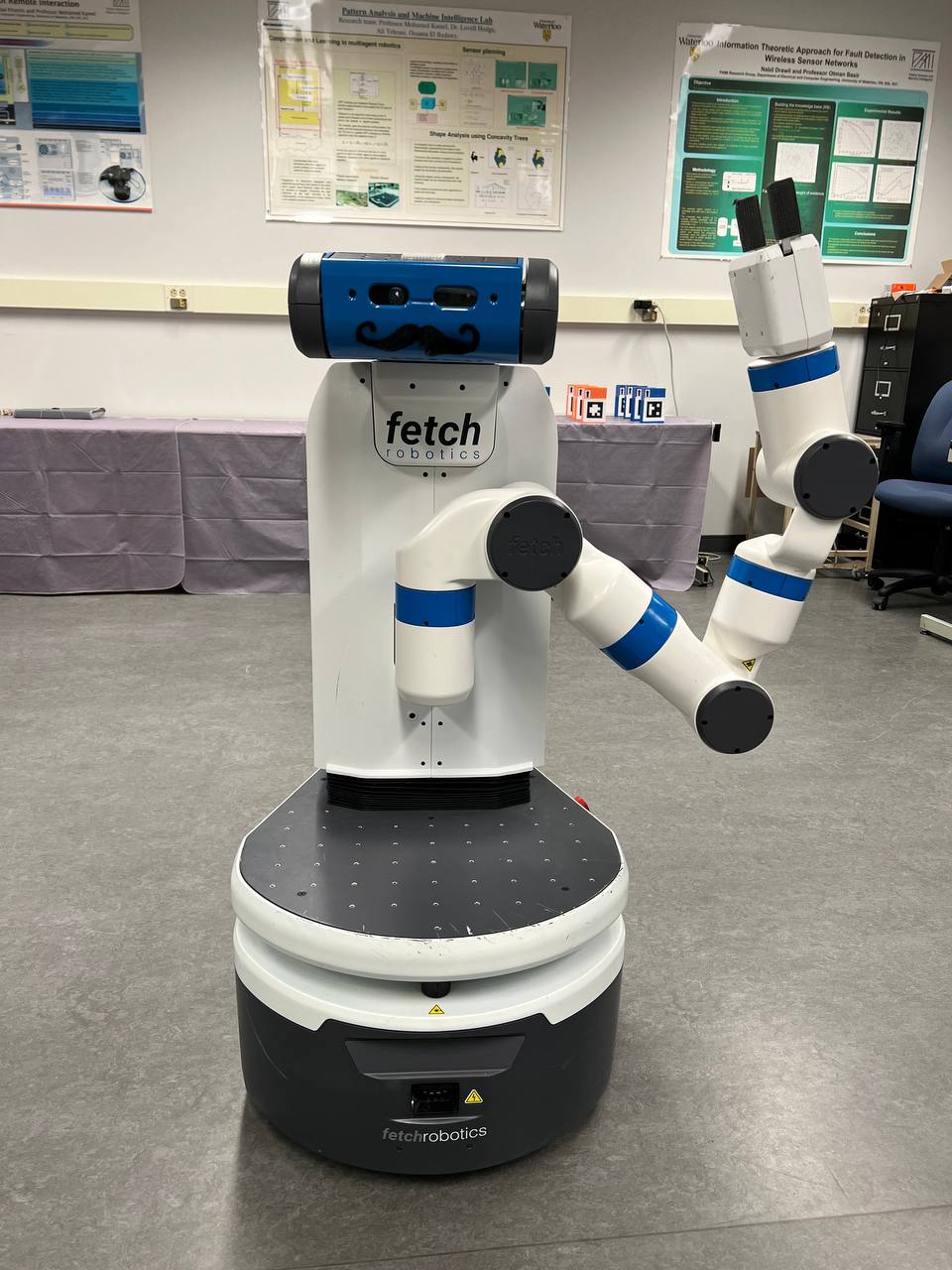}
      }
    \hfill
    \subfloat[Table 2: Blue and pink objects are on this table\label{fig:d3}]{%
        \includegraphics[width=0.23\textwidth]{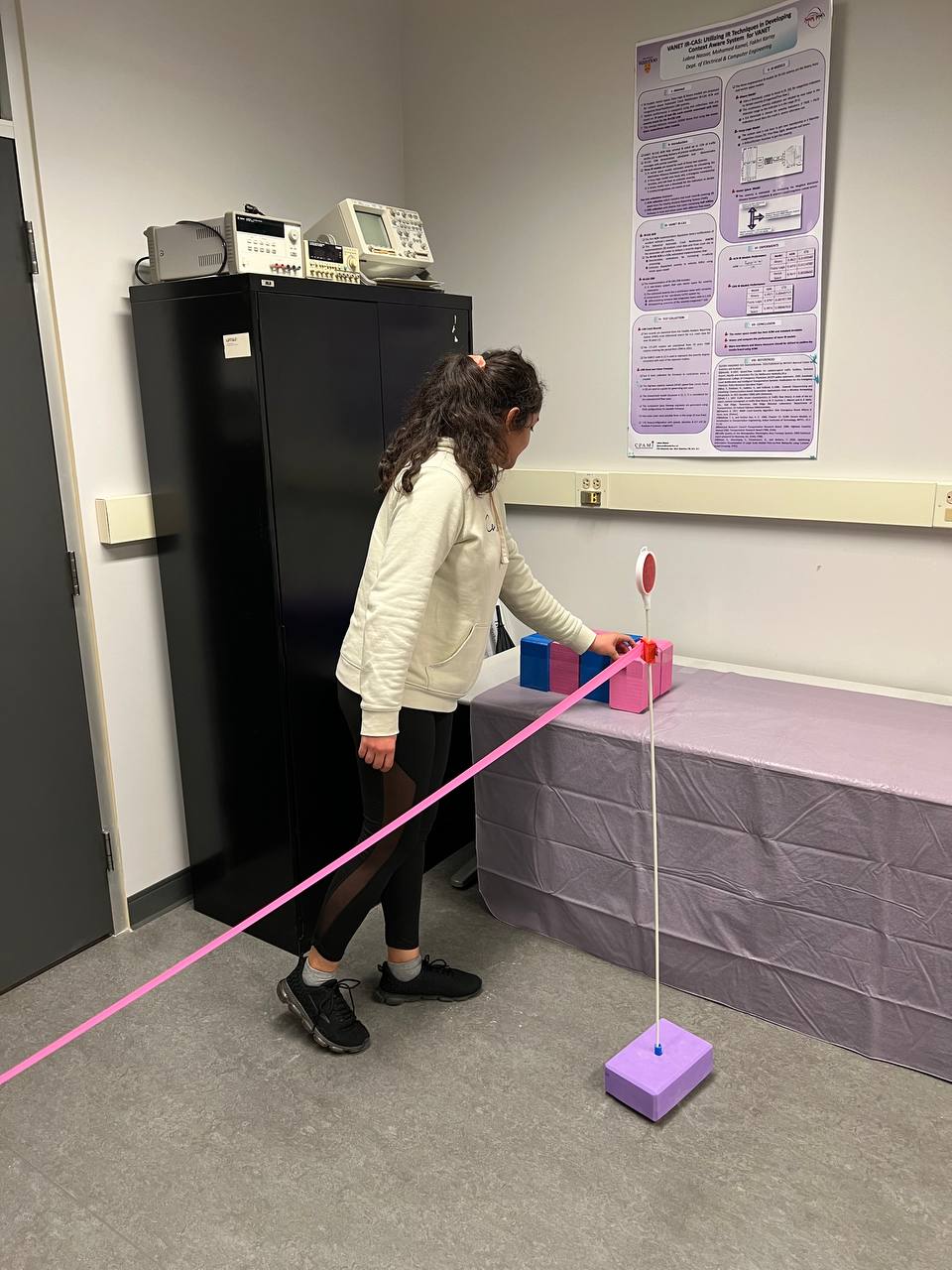}
      }
    \hfill
    \subfloat[The human is placing an object on the shared workspace\label{fig:d4}]{%
        \includegraphics[width=0.23\textwidth]{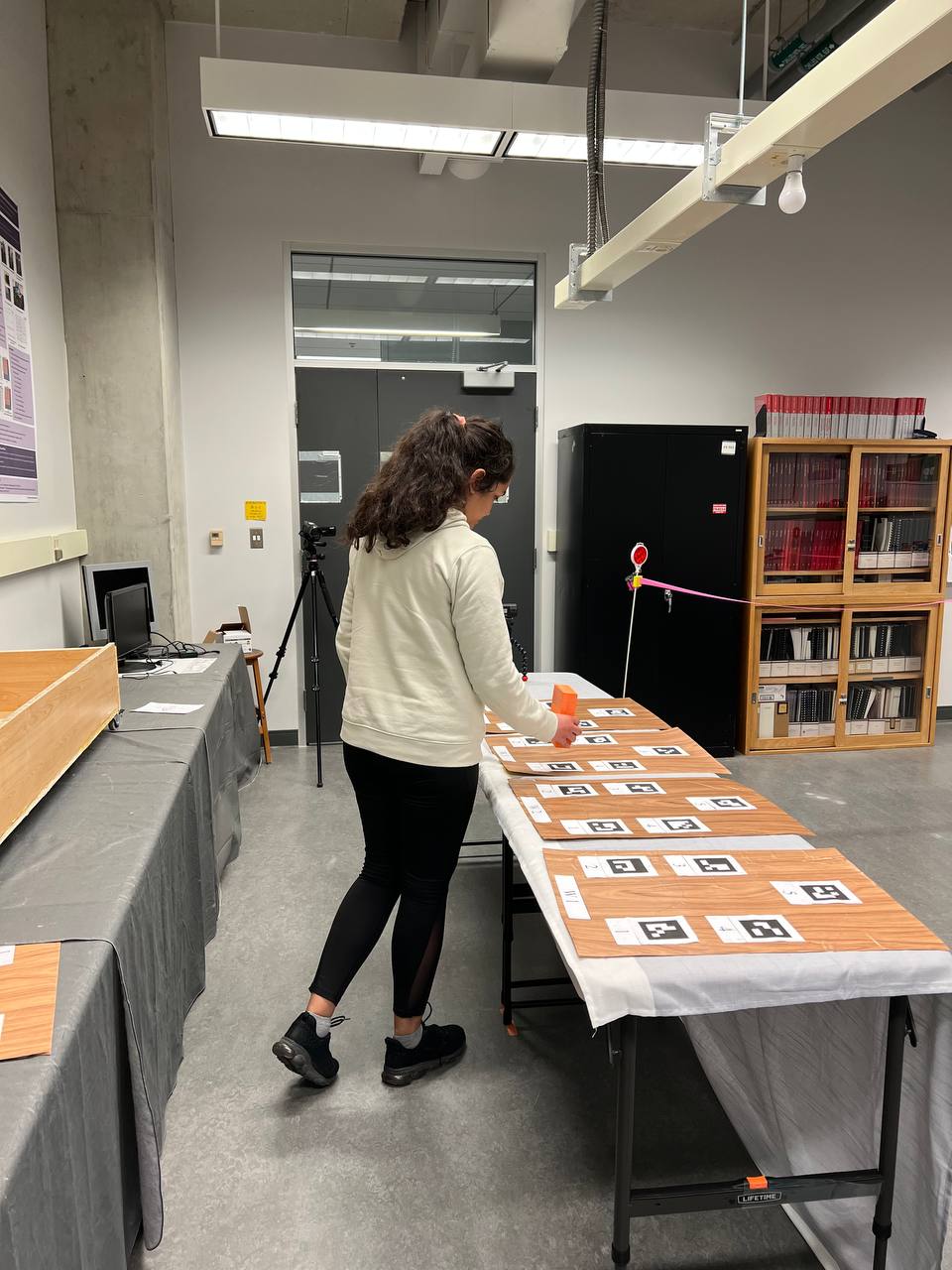}
      }
    \caption{Scenes from the human-robot collaboration scenario used for system evaluation.}
    \label{fig:exp_begin}
\end{figure*}

Fig.~\ref{fig:exp_setup} shows the experimental setup consisting of tables for the human and robot, as well as a shared workspace (matching Fig.~\ref{fig:exp_outline}). The experiment begins with the human at Table 1 (Fig.~\ref{fig:d1}), where the tablet is located. The robot stays put (Fig.~\ref{fig:d2}) until the human selects their task and possibly assigns some tasks to the robot through the GUI. This helps the robot better estimate the human's leading/following preference. Next, the human can retrieve their selected object from Tables 1 or 2 (Fig.~\ref{fig:d1} and \ref{fig:d3}) and place it on the chosen workspace and spot (Fig.~\ref{fig:d4}). To ensure safety, a light bulb is installed on top of the shared workspace, which turns red when the robot approaches, indicating that the human should avoid getting too close to the shared workspace.

\section{System Validation Study}\label{sec:4}
To evaluate the planning framework, we analyzed the robot's decisions for four distinct behaviours emulated (enacted) by the experimenter. Before detailing the results of the system evaluation, we will explain some preliminaries to aid comprehension.

\textbf{Collaboration task}: The tasks and their corresponding precedence constraints are represented by a directed acyclic graph known as a task graph. 

\textbf{Human following preference}: 
We model the human as having a hidden preference to lead or follow in the interaction, given as a scalar quantity $p_f$. We treat this as a random variable $p_f$ taking values in the discrete set $V = \big(0.0, 0.2, 0.4, 0.6, 0.8, 1\big)$. The values $p_f=0$ and $p_f=1$ respectively mean that the human very likely prefers to lead and follow the robot. At the beginning of the experiment, the robot assumes that the human will follow it, and sets its initial belief about that as follows:
\begin{align*}
    &P[p_f=v_i] = b(i; n=5, p=0.8), \; V=(v_0, \dots, v_5),
\end{align*}
where $b(i;n,p)$ is a binomial distribution.

\textbf{Human Performance}: Similar to human preferences, we consider human accuracy as a latent variable in the model of human performance. We capture this variable using a single scalar random variable denoted by $p_e$, which takes on values from a tuple of discrete values $W={0, 0.1, 0.2, ..., 1}$. A value of $p_e=1$ represents a high likelihood of errors by the human, while $p_e=0$ indicates accuracy. The robot starts with the assumption that the human is mostly accurate and updates this belief, represented by a binomial distribution, during the collaboration:
\begin{align*}
    &P[p_e=w_i] = b(i; n=10, p=0.1), \; W=(w_0, \dots, w_{10}).
\end{align*}

\textbf{Task allocation}: 
The task allocation problem is modeled as a cost function, where penalties are assigned for specific actions. A penalty is incurred for assigning a task to a human who prefers to lead, and for not assigning tasks to the human who is prone to making mistakes. By assigning tasks to the human, the robot informs them about the next object that needs to be placed in a workspace. This can limit the human's autonomy, but it also helps to prevent mistakes. Additionally, there is a penalty for assigning a task to the robot that has already been assigned to a human.


\textbf{Updating $p_e$ and $p_f$}: 
To update the robot's belief regarding the human's performance level and inclination to follow the robot, we have developed models that capture the transition between various levels of accuracy and preference, alongside models for human accuracy and preference. The belief of the robot is then updated through the utilization of the approach presented in \cite{nikolaidis2017human}. Further information on the models and methodology can be found in \cite{noormohammadi2022task}.

\subsection{System Evaluation Results: Enacting Different Human Behaviours}
In this part, we evaluated the effectiveness of the proposed framework by conducting experiments that enacted four distinct types of human behaviour. 

Results are shown in Table~\ref{tbl:results}, where $n^{\textit{wrong}}_h$, $n_h^\textit{assign}$,  and $n_r^\textit{assign}$ are the number of the  human's wrong actions, the number of actions assigned to the robot by the human, and the number of actions assigned to the human by the robot. In addition, we measured the total number of actions done by the human and robot, denoted respectively by $n_h^T$ and $n_r^T$. The travel distance by the robot and human were also measured approximately by measuring the distance between the picking and placing tables. The last column of the table, $t_{\text{tot}}$, shows the task completion time.
\begin{table*}[tbh]
\caption{Evaluation results: Four different enacted behaviours (one run per case) }
\label{tbl:results}
\newcolumntype{Y}{>{\centering\arraybackslash}X}
\begin{tabularx}{\textwidth}{YY|c|YY|YY|YY|YY|c}
 \multicolumn{2}{c}{Human} & \multicolumn{1}{c}{Wrong actions} & \multicolumn{2}{c}{Assigned tasks} & \multicolumn{2}{c}{Travel distance (m)} & \multicolumn{2}{c}{Total Tasks} & \multicolumn{1}{c}{Total time (min)} \\ \toprule

\multicolumn{1}{Y}{Preference} & \multicolumn{1}{Y|}{Accuracy} 
& \multicolumn{1}{c|}{$n^{\textit{wrong}}_h$}  
& \multicolumn{1}{Y}{$n_h^\textit{assign}$}  & \multicolumn{1}{Y|}{$n_r^\textit{assign}$}  

& \multicolumn{1}{Y}{$d_h^T$} & \multicolumn{1}{Y|}{$d_r^R$} 
& \multicolumn{1}{Y}{$n_h^T$}  & \multicolumn{1}{Y|}{$n_r^T$}  
& \multicolumn{1}{c}{$t_{\textit{total}}$} \\\midrule[1pt]

Follow & High & 
0 & 0 & 9 & 286 & 36 & 13 & 7 & \multicolumn{1}{c}{7.65}  \\
Follow & Low & 
2 & 0 & 11 & 300 & 48 & 14 & 10 & \multicolumn{1}{c}{10.1} \\
Lead & Low & 
 8 & 7 & 12 & 300 & 64 & 14 & 10 & \multicolumn{1}{c}{13.1}\\
Lead & High & 
 0 & 10 & 0 & 180 & 76 & 11 & 9 & \multicolumn{1}{c}{11.1} \\
 \bottomrule[1.3pt]
\end{tabularx}
\end{table*}

\begin{figure}[t]
      \centering
      \subfloat[Following preference with high accuracy\label{fig:fh}]{%
        \includegraphics[scale=0.26]{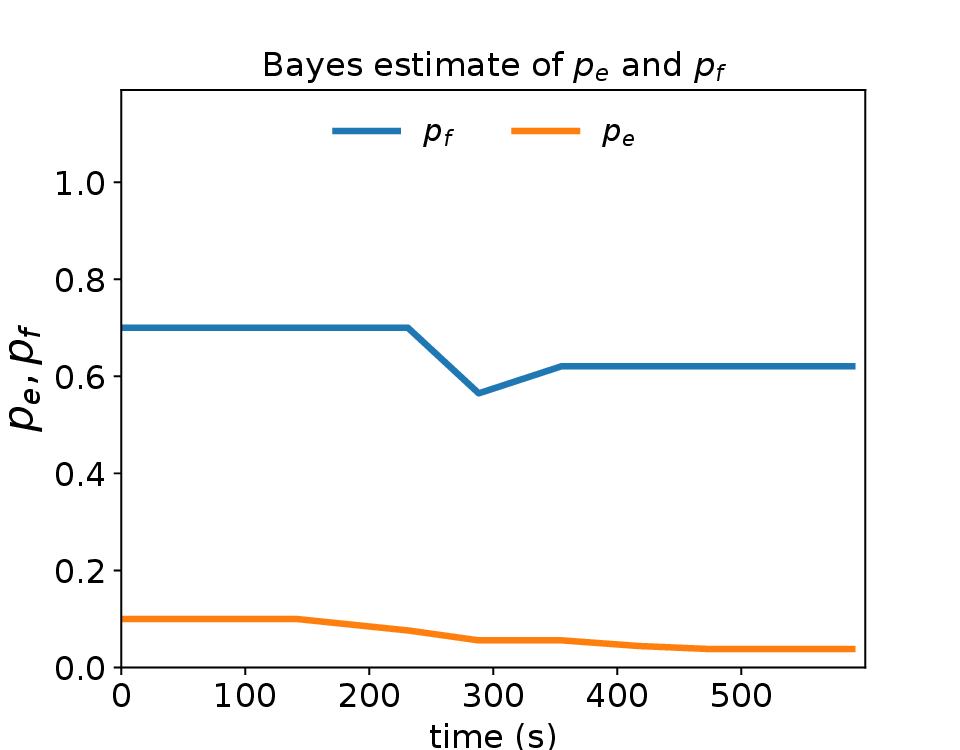}
      }
      \hfill
      \subfloat[Following preference with low accuracy\label{fig:fl}]{%
        \includegraphics[scale=0.26]{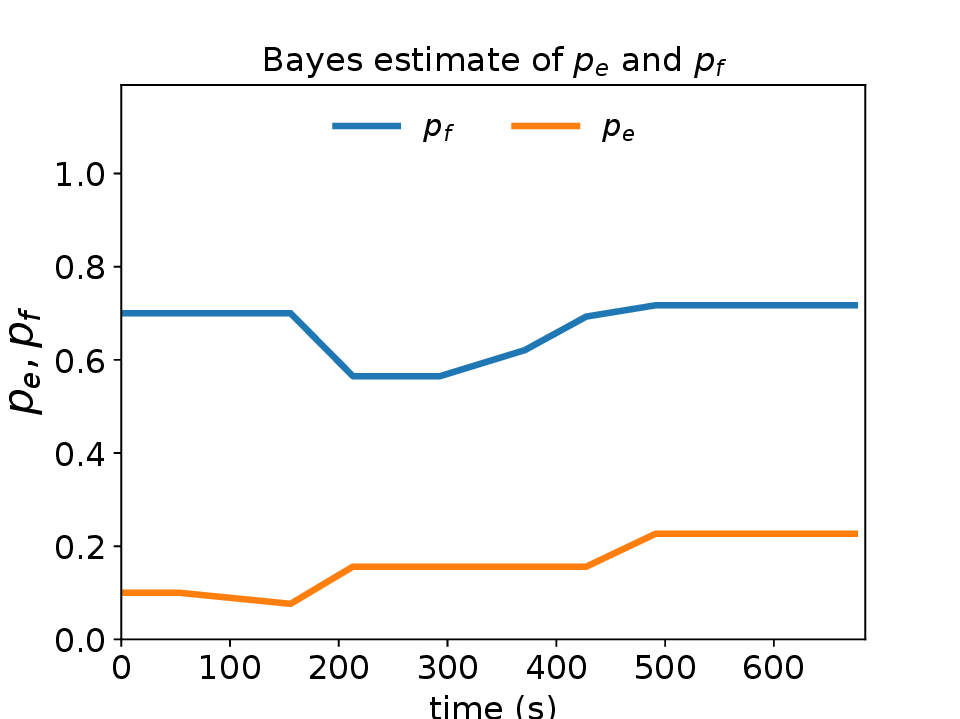}%
      }
    \hfill
      \subfloat[Leading preference with low accuracy\label{fig:ll}]{%
        \includegraphics[scale=0.26]{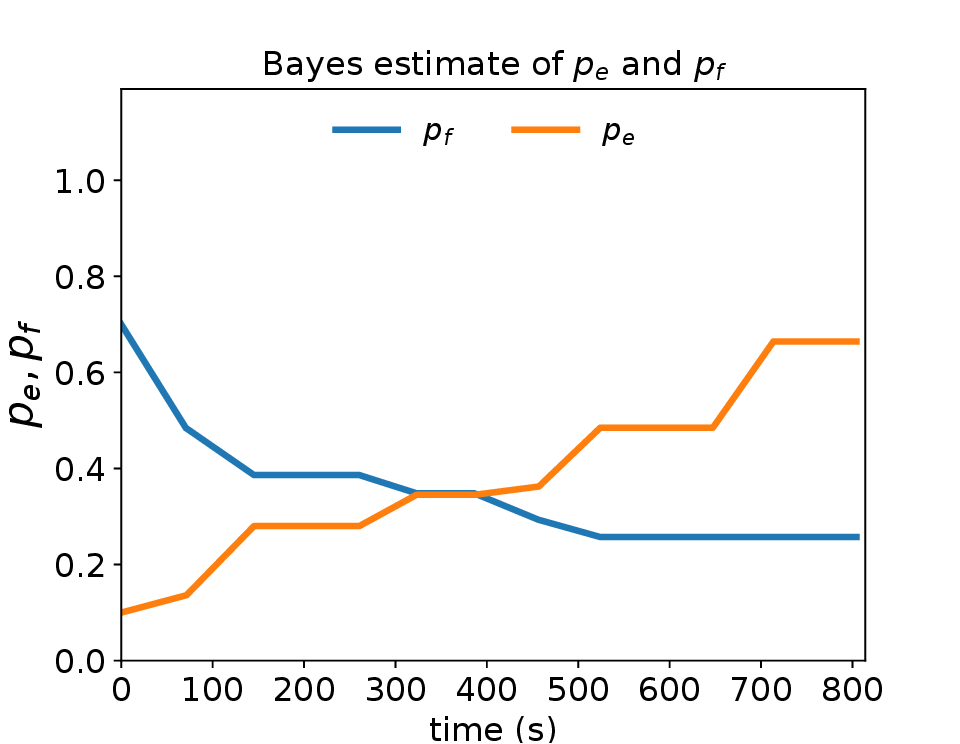}%
      }
     \hfill
      \subfloat[Leading preference with high accuracy\label{fig:lh}]{%
        \includegraphics[scale=0.26]{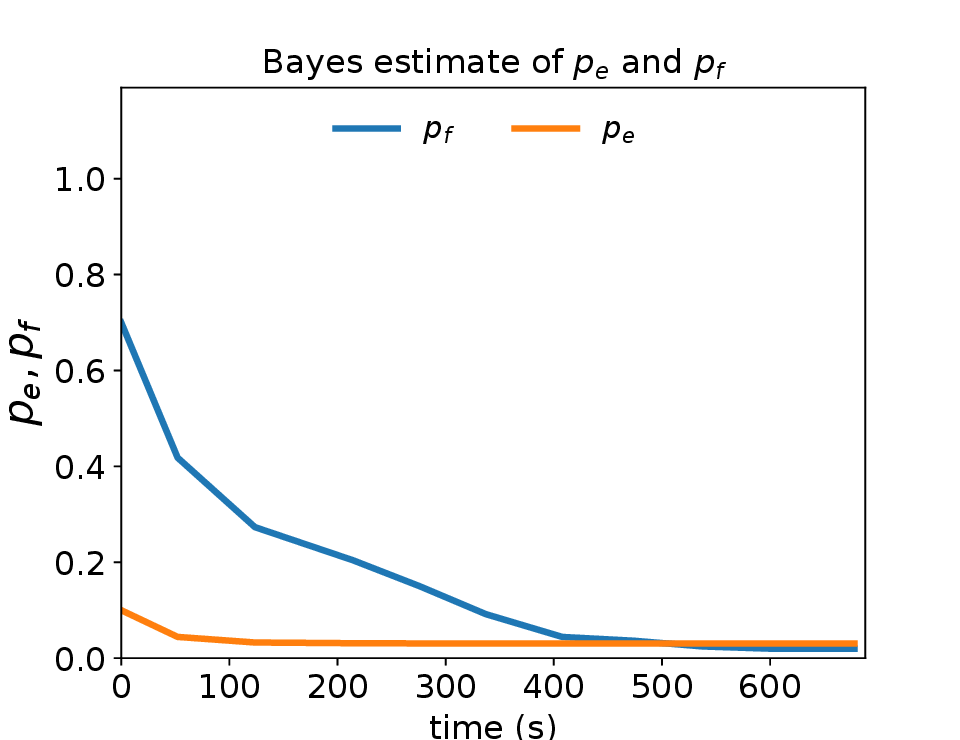}%
      }
      \caption{The robot's belief regarding the human's preference and performance (computed as the expectation of the posterior distribution)}
      \label{fig:lflh}
\end{figure}

Based on Table~\ref{tbl:results}, as expected, the team has the best performance in terms of collaboration time when the human preferred to follow the robot and had high accuracy. In this case, the human did not have any mistakes and followed the optimal plan generated by the robot. The robot assigned more tasks related to blue objects (which were far from both the human and the robot) to the human agent since they could complete those tasks more quickly. The robot also gave the human agent more tasks involving orange objects, which were closer to the human and farther from the robot. Meanwhile, the robot handled the tasks with pink objects because they were nearer to the robot and distant from the human agent. As a result, the human agent traveled a greater distance than the robot, since they completed more tasks that were farther away.

In the second scenario, where the human had low accuracy and had the potential to make mistakes, the team's performance was still somewhat similar to the previous scenario since the human followed the robot and didn't make any mistakes. In other words, despite the human's potential to make errors, following the robot's plan prevented the human from making any mistakes. However, a few mistakes by the human agent caused the robot to assign more tasks to it, and the collaboration time increased as the robot needed to fix the human's errors. 

In the scenario where the human preferred to lead but had low accuracy, more tasks were assigned to the robot, and most of them were wrong. Additionally, the human agent placed some wrong objects on the shared workspace. The robot was able to recognize the human's low accuracy and gradually assumed more responsibility for leading and assigning tasks. However, the human's preference to lead, combined with their low accuracy, resulted in a longer completion time and greater travel distance compared to the other scenarios. 
In the case where the human preferred to lead and had high accuracy, the robot allowed the human to take the leading role. However, the human agent assigned more tasks related to blue objects (which were far from both the human and the robot) to the robot. As a result, the robot traveled a greater distance, while the human agent traveled less, leading to a longer overall completion time.

Fig.~\ref{fig:lflh} shows the changes in the robot's belief regarding the human agent's leading preference and accuracy in each scenario. In Figures \ref{fig:fh} and \ref{fig:fl}, the robot estimated that the human preferred to follow. In the second scenario, it was assumed that the human had low accuracy and forgot the pattern. However, in both the first and second scenarios, the human agent followed the robot without making significant mistakes, leading the robot to consider them as nearly accurate. As depicted in Fig.~\ref{fig:fh} and \ref{fig:fl}, the robot perceived the human as an accurate agent, despite its accuracy estimation being slightly lower in the second scenario compared to the first. In Figures \ref{fig:ll} and \ref{fig:lh}, the robot accurately estimated  the human agent's both leading preference and accuracy, allowing it to plan accordingly.

\begin{figure}
    \centering
    \includegraphics[width=0.4\textwidth]{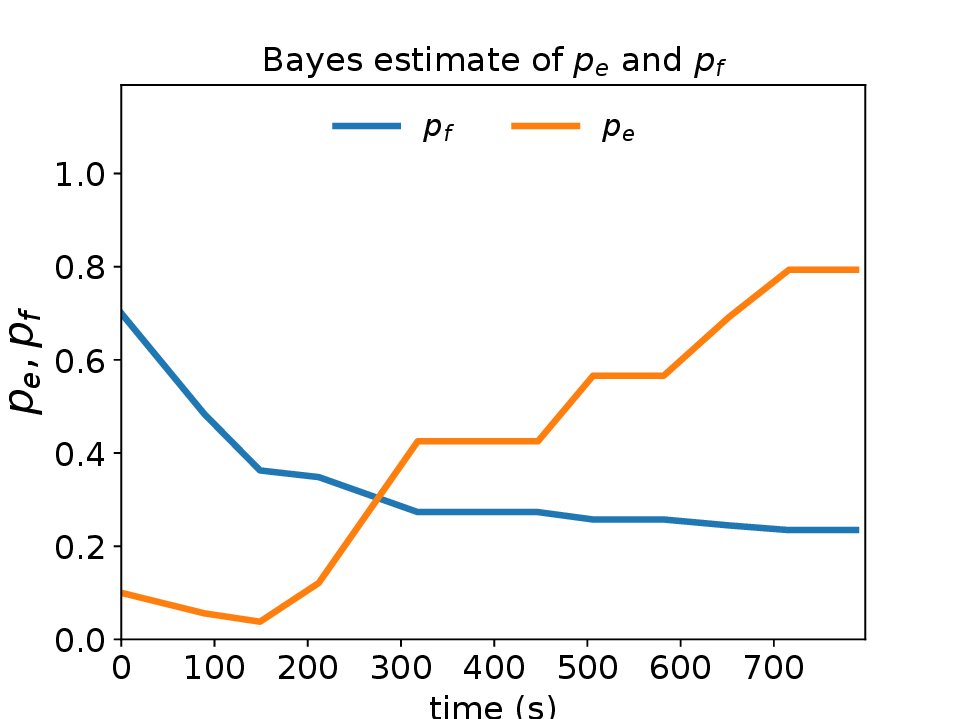}
    \caption{Robot's belief regarding the human's performance and preference. The human first had high accuracy but then made more errors}
    \label{fig:varying_perf}
\end{figure}

In addition to the four different cases previously discussed, we also examined a scenario where the human initially performed well but later began to make errors. The video of this scenario is available\footnote{https://youtu.be/gbyDUV0ZLfI}. Fig.~\ref{fig:varying_perf} illustrates the evolution of the robot's belief regarding the human's performance and preference throughout the experiment. The robot's belief about the human's preference decreased consistently since the human wanted to be the leader throughout the experiment. However, the robot's belief about the human's performance initially decreased, indicating that the human was accurate, but later increased as the human made more errors.





\section{CONCLUSION}

In this paper, we focused on the implementation of the proposed task planning framework for human-robot collaboration in a real scenario. Our proposed framework offers a novel approach that considers human preference and performance, and our previous study showed the effectiveness of the proposed framework in a simulation environment. However, to assess its efficiency, it was necessary to test it in a real environment with possible arising uncertainties.  The successful implementation of the algorithm with the Fetch robot in a real-world scenario showed that our planning framework has enabled the robot to adapt itself based on different human preferences and performances (enacted by the experimenter).  Our framework has the potential to be applied in various settings and could contribute to the development of improved human-robot collaboration. In this paper, we only considered four different variations of human behaviour, including different combinations of leading/following preference and high/low accuracy, emulated by the experimenter. However, in future research, we will conduct a user study and evaluate the planning framework with  participants working in a team with the Fetch robot. Moreover, we will investigate the impact of different factors, such as the human's leadership/ followership styles or task difficulty, on collaboration and the human perception of the robot.





\bibliographystyle{IEEEtran}
\bibliography{main}

\end{document}